\documentclass[conference]{IEEEtran}
\IEEEoverridecommandlockouts
\usepackage{cite}
\usepackage{amsmath,amssymb,amsfonts}
\usepackage{algorithmic}
\usepackage{graphicx}
\usepackage{textcomp}
\usepackage{xcolor}
\usepackage{float}
\usepackage{hyperref}
\def\BibTeX{{\rm B\kern-.05em{\sc i\kern-.025em b}\kern-.08em
    T\kern-.1667em\lower.7ex\hbox{E}\kern-.125emX}}
\begin{document}

\title{YOLOv5-Based Object Detection for Emergency Response in Aerial Imagery}
\author{
    \IEEEauthorblockN{Sindhu Boddu}
    \IEEEauthorblockA{
        \textit{Department of Electrical and Computer} \\
		\textit{Engineering} \\        
        \textit{UNC Charlotte}\\
        Charlotte, North Carolina, USA \\
        sboddu2@charlotte.edu}
    \and
    \IEEEauthorblockN{Dr. Arindam Mukherjee}
    \IEEEauthorblockA{
        \textit{Director of Cyber-Physical Systems Lab} \\
        \textit{Department of Electrical and Computer } \\
        \textit{Engineering} \\ 
        \textit{UNC Charlotte}\\
        Charlotte, North Carolina, USA \\
        amukherj@charlotte.edu}
   
}

\maketitle

\begin{abstract}
This paper presents a robust approach for object detection in aerial imagery using the YOLOv5 model. We focus on identifying critical objects such as ambulances, car crashes, police vehicles, tow trucks, fire engines, overturned cars, and vehicles on fire. By leveraging a custom dataset, we outline the complete pipeline from data collection and annotation to model training and evaluation. Our results demonstrate that YOLOv5 effectively balances speed and accuracy, making it suitable for real-time emergency response applications. This work addresses key challenges in aerial imagery, including small object detection and complex backgrounds, and provides insights for future research in automated emergency response systems.
\end{abstract}


\section{Introduction}
Object detection in aerial imagery is a vital component of modern surveillance and disaster response systems. It facilitates rapid assessment of emergency scenarios, enabling timely decision-making. The YOLOv5 model, known for its high accuracy and real-time capabilities, is particularly suited for such tasks. This paper focuses on detecting critical objects in aerial images, including ambulances, police vehicles, and fire trucks, to assist emergency responders.

Aerial imagery poses unique challenges, such as detecting small and occluded objects against complex backgrounds. This study addresses these challenges by employing a customized YOLOv5 model and a dataset tailored to emergency scenarios. The main objectives are to train a robust detection model and evaluate its performance under real-world conditions.

This paper is organized as follows: Section II reviews related work, discussing traditional methods like Histogram of Oriented Gradients (HOG) and Support Vector Machines (SVM) alongside modern deep learning models. Section III describes the dataset preparation process, including data collection, annotation, and splitting. Section IV introduces the YOLOv5 model architecture, detailing the modifications made for aerial imagery. Section V outlines the methodology, including preprocessing steps, training setup, and evaluation metrics. Section VI presents the results, covering validation metrics, test performance, and class-wise analysis. Section VII highlights challenges and insights gained during the study. Section VIII compares YOLOv5 with other object detection models, and Section IX explores potential real-world applications. Finally, Section X concludes the study with key findings and directions for future research.
\section{Related Work}
Object detection has evolved significantly with the advent of deep learning. Traditional methods, such as Histogram of Oriented Gradients (HOG) and Support Vector Machines (SVM), have been widely used for feature extraction and classification in earlier object detection systems. However, these methods rely heavily on handcrafted features and struggle with complex scenes. Convolutional Neural Networks (CNNs) have replaced these traditional approaches, enabling models like Faster Region-based CNN (Faster R-CNN), Single Shot Detector (SSD), and You Only Look Once (YOLO) to achieve significant advancements. Among these, the YOLO family of models revolutionized real-time detection with its end-to-end pipeline, and YOLOv5 offers improved accuracy and flexibility.

Despite extensive research, few studies focus on emergency-related object detection in aerial imagery. Existing works address general object detection but lack specialization in scenarios requiring rapid response. This paper bridges this gap by targeting emergency-specific objects in aerial views.
\section{Data Set}
\subsection{Data Collection}

A custom dataset was created by sourcing aerial images from drones and public repositories. The dataset includes diverse scenarios such as accidents, fires, and emergency vehicle presence, ensuring a wide range of conditions.
\subsection{Classes of Interest}
The dataset focuses on the following classes:
\begin{itemize}
    \item Ambulance
    \item Car crash
    \item Police vehicle
    \item Tow truck
    \item Fire engine
    \item Car upside down
    \item Car on fire
\end{itemize}
\subsection{Annotation}
Images were annotated using the LabelImg tool, saving labels in YOLO format. Each label file specifies the class and bounding box coordinates. Careful annotation ensures accurate training and evaluation.

\subsection{Data Split}
The dataset was divided as follows:
\begin{itemize}
    \item 70\% for training
    \item 15\% for validation
    \item 15\% for testing
\end{itemize}
This split ensures sufficient data for training while reserving enough for performance evaluation.
\section{YOLOv5 Model Architecture}
YOLOv5 is a one-stage object detector known for its real-time detection capabilities. It uses a Cross-Stage Partial Network (CSPNet) backbone for feature extraction, a Path Aggregation Network (PANet) neck for feature aggregation, and YOLO heads for final predictions. These components work together to enable efficient and accurate object detection.
Modifications to the architecture include:
\begin{itemize}
    \item Adjusting anchor sizes for aerial imagery.
    \item Increasing input resolution to improve small object detection.
\end{itemize}
These enhancements enable YOLOv5 to handle the unique challenges of aerial images effectively.

\section{Methodology}
\subsection{Preprocessing}
Images were resized and normalized. Data augmentation techniques, such as flipping, rotation, and color jitter, were applied to improve generalization.

\subsection{Training Setup}
\begin{itemize}
    \item \textbf{Model:} YOLOv5 small (yolov5s.pt)
    \item \textbf{Batch Size:} 16
    \item \textbf{Epochs:} 100
    \item \textbf{Image Size:} 640x640
    \item \textbf{Hardware:} Deep Learning Box (NVIDIA GTX 1080 Ti)
\end{itemize}

\subsection{Evaluation Metrics}
The model was evaluated using several key metrics to comprehensively assess its performance on the object detection task:

\begin{itemize}
    \item \textbf{mAP (mean Average Precision)}:
    \begin{itemize}
        \item \textbf{mAP@0.5}: This metric calculates the mean Average Precision at an Intersection over Union (IoU) threshold of 0.5. It is widely used in object detection tasks as it provides a balance between precision and recall, emphasizing the model's ability to detect objects with a reasonable overlap with ground truth bounding boxes. In the context of aerial imagery, mAP@0.5 is crucial as it highlights the model's accuracy in localizing objects such as ambulances and fire engines.
        \item \textbf{mAP@0.5:0.95}: This metric computes the mean Average Precision across multiple IoU thresholds ranging from 0.5 to 0.95 in increments of 0.05. Unlike mAP@0.5, which focuses on moderate overlap, mAP@0.5:0.95 provides a more stringent evaluation, testing the model's robustness in localizing objects precisely. This metric is particularly significant in scenarios like detecting small or partially occluded objects in aerial imagery, where higher IoU thresholds demand precise bounding box predictions.
    \end{itemize}

    \item \textbf{Precision}: Precision measures the percentage of correctly predicted objects among all detections. High precision indicates fewer false positives, which is critical in emergency response scenarios where false alarms can lead to unnecessary actions.
    
    \item \textbf{Recall}: Recall quantifies the proportion of true objects correctly identified by the model. In the context of aerial imagery, high recall ensures that most emergency-related objects (e.g., tow trucks or police vehicles) are detected, minimizing the risk of missing crucial information.
    
    \item \textbf{F1-Score}: This harmonic mean of precision and recall provides a balanced measure of the model's performance, especially when precision and recall are in trade-off.
    
    \item \textbf{IoU (Intersection over Union)}: IoU evaluates the overlap between predicted bounding boxes and ground truth boxes. This metric is foundational for calculating mAP and ensures that object localization meets the necessary thresholds for accurate detection.
\end{itemize}

These metrics were chosen to comprehensively evaluate the YOLOv5 model's performance, ensuring it is well-suited for real-time emergency response applications. mAP@0.5 serves as a standard benchmark, while mAP@0.5:0.95 challenges the model to achieve higher precision in more complex scenarios. The combination of these metrics ensures the robustness and reliability of the model across diverse detection tasks.

\section{Results}
\subsection{Dataset Details}
The dataset consists of the following:
\begin{itemize}
    \item \textbf{Classes:} Ambulance, Car crash, Car upside down, Tow truck, Fire engine, Police, Car on fire
    \item \textbf{Total Images:} 772
    \item \textbf{Train/Val/Test Split:} 70\% training, 15\% validation, 15\% testing
\end{itemize}

\subsection{Validation Results}
Key metrics on the validation set are as follows:
\begin{itemize}
    \item \textbf{mAP (IoU=0.5):} 46.7\%
    \item \textbf{mAP (IoU=0.5:0.95):} 27.9\%
    \item \textbf{Precision:} 49.6\%
    \item \textbf{Recall:} 43.1\%
    \item \textbf{F1 Score:} 46.1\%
\end{itemize}

\subsection{Test Results}

Visualizations of the test images demonstrate the model's strengths and weaknesses across different classes.

\begin{figure}[H] 
    \centering
    \includegraphics[width=\linewidth]{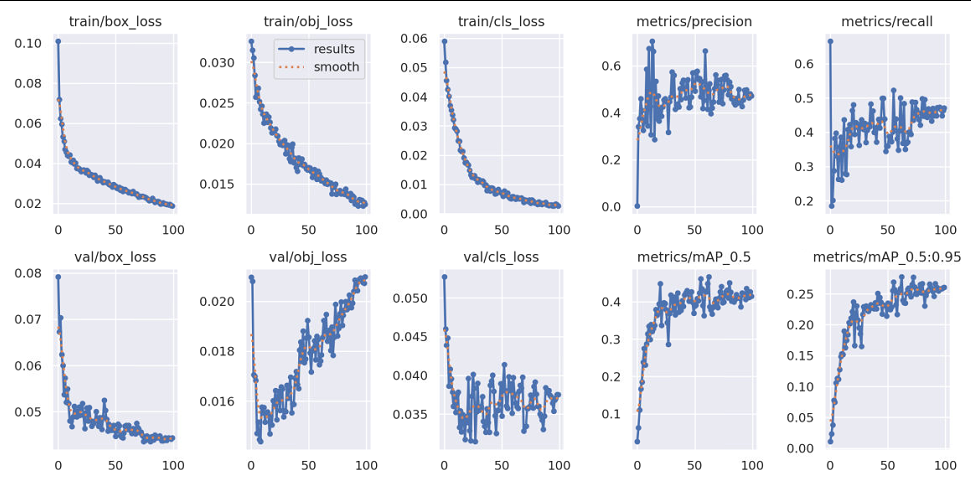}
    \caption{Training and validation metrics of the YOLOv5 model over 100 epochs.}
    \label{fig:results1}
\end{figure}
The figure illustrates the training and validation performance of the YOLOv5 model over 100 epochs. Training losses (\texttt{train/box\_loss}, \texttt{train/obj\_loss}, \texttt{train/cls\_loss}) show a consistent decline, indicating improved localization, objectness, and classification accuracy. Validation losses (\texttt{val/box\_loss}, \texttt{val/obj\_loss}, \texttt{val/cls\_loss}) follow a similar trend with minor fluctuations, reflecting challenges in generalization. Performance metrics, including precision, recall, mAP@0.5, and mAP@0.5:0.95, demonstrate steady improvement. Precision reflects detection accuracy, while mAP metrics highlight overall performance, with mAP@0.5:0.95 evaluating robustness across IoU thresholds. The results confirm the model’s convergence, achieving a balance between loss minimization and detection performance.
\begin{figure}[H]
    \centering
    \includegraphics[width=\linewidth]{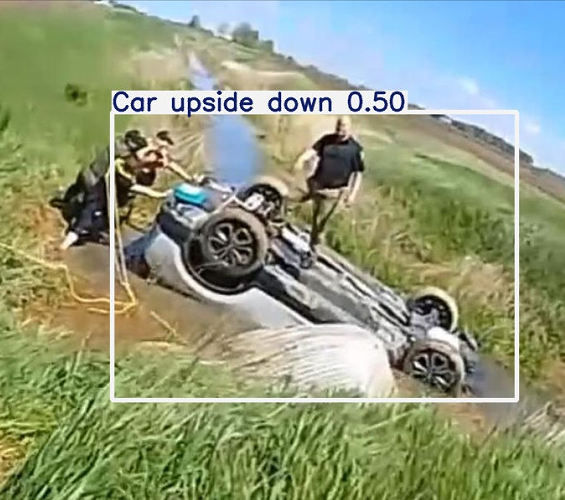}
    \caption{The model detects a flipped car with 0.50 confidence, highlighting challenges in distinguishing rare objects due to background complexity and irregular shapes.}
    \label{fig:car_upside_down}
    \end{figure}

\begin{figure}[H]
    \centering
    \includegraphics[width=\linewidth]{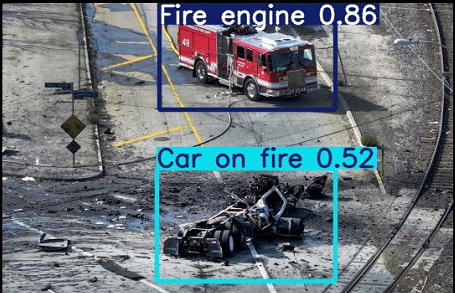}
    \caption{The model detects a fire engine (0.86 confidence) and a car on fire (0.52 confidence), showcasing strong performance for distinct objects but struggles with smaller, visually complex ones.}
    \label{fig:fire_engine_car_on_fire}
    \end{figure}

\begin{figure}[H]
    \centering
    \includegraphics[width=\linewidth]{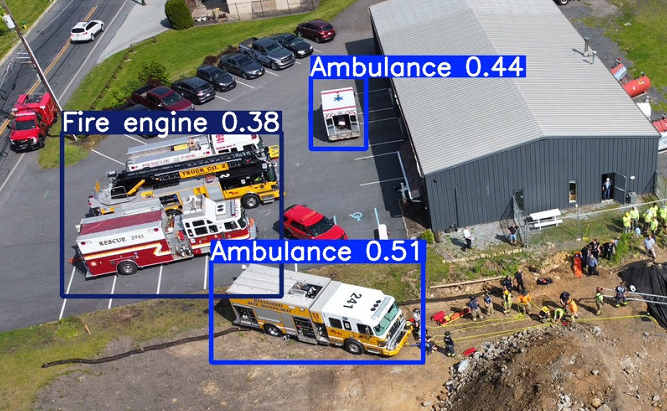}
      \caption{The model detects two ambulances (0.51, 0.44) and a fire engine (0.38), highlighting challenges with occlusion and complex backgrounds.}
    \label{fig:results4}
\end{figure}

\begin{figure}[H]
    \centering
    \includegraphics[width=\linewidth]{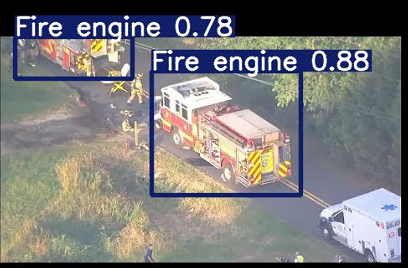}
    \caption{The model detects two fire engines with confidence scores of 0.88 and 0.78, demonstrating strong performance for large, distinct objects.}
    \label{fig:results5}
\end{figure}

\begin{figure}[H]
    \centering
    \includegraphics[width=\linewidth]{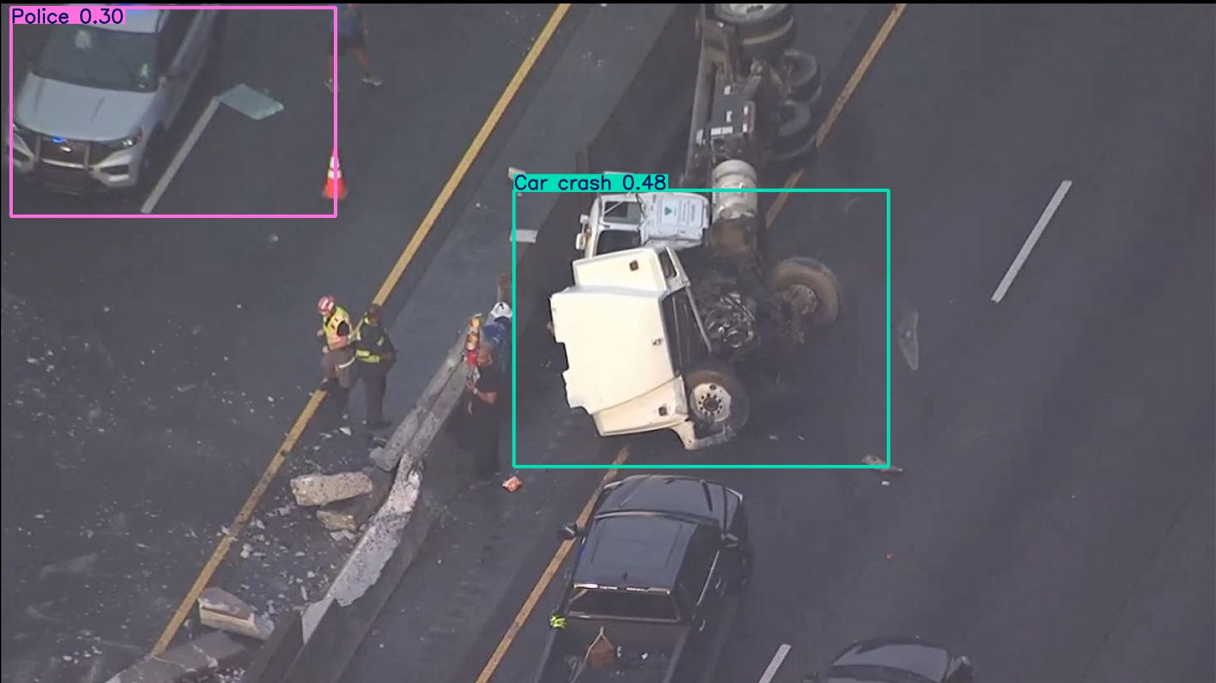}
    \caption{The model detects a car crash (0.48 confidence) and a police vehicle (0.30 confidence), showcasing challenges with low confidence in cluttered environments.}
    \label{fig:results6}
\end{figure}

\begin{figure}[H]
    \centering
    \includegraphics[width=\linewidth]{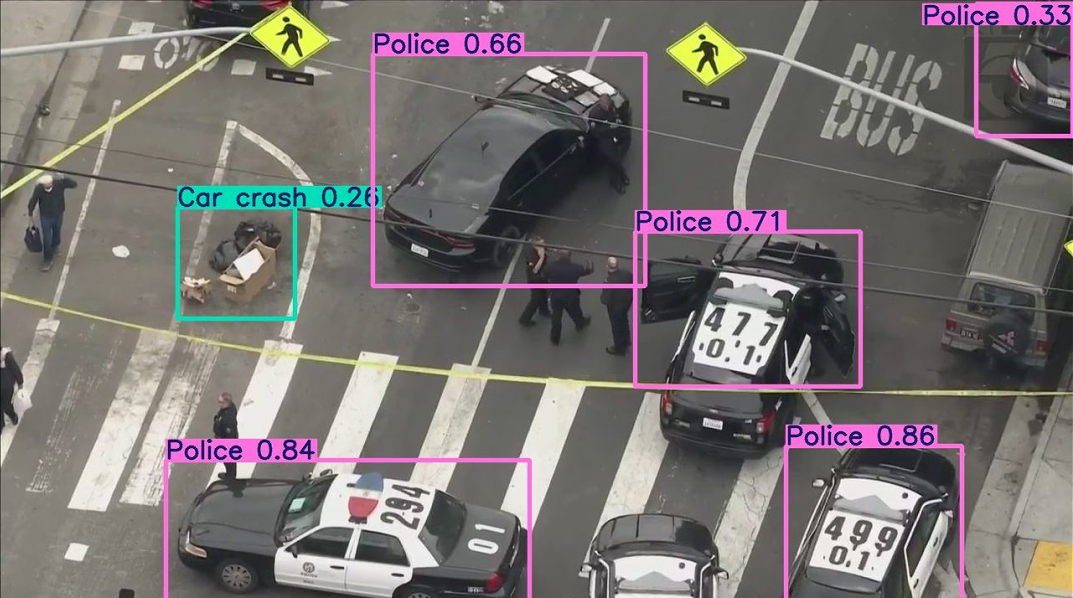}
    \caption{The model detects multiple police vehicles (confidence: 0.33–0.86) and a car crash (confidence: 0.28), showcasing strong performance for distinct objects but challenges with smaller detections.}
    \label{fig:results7}
\end{figure}

\begin{figure}[H]
    \centering
    \includegraphics[width=\linewidth]{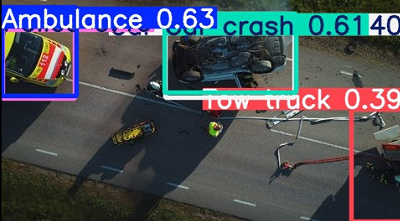}
    \caption{The model detects an ambulance (0.63), a car crash (0.61), and a tow truck (0.39), demonstrating accurate detection for prominent objects but lower confidence for smaller ones.}
    \label{fig:results8}
\end{figure}

\subsection{Class-Wise Performance}
\begin{table}[ht]
\caption{Class-Wise Performance Metrics}
\label{tab:class_perf}
\centering
\begin{tabular}{|l|c|c|c|c|c|}
\hline
\textbf{Class} & \textbf{Precision} & \textbf{Recall} & \textbf{mAP@50} & \textbf{mAP@50-95} & \textbf{F1 Score} \\ \hline
Ambulance      & 67.3\%            & 72.2\%         & 58.8\%          & 43.6\%             & 69.7\%           \\ \hline
Car on fire    & 18.1\%            & 7.1\%          & 11.9\%          & 6.04\%             & 10.3\%           \\ \hline
Car upside down& 17.5\%            & 11.1\%         & 14.6\%          & 12.2\%             & 13.7\%           \\ \hline
Car crash      & 46.1\%            & 27.6\%         & 36.7\%          & 15.2\%             & 34.6\%           \\ \hline
Fire engine    & 61.5\%            & 37.9\%         & 51.3\%          & 34.2\%             & 47.0\%           \\ \hline
Police         & 56.2\%            & 74.2\%         & 70.1\%          & 47.9\%             & 63.9\%           \\ \hline
Tow truck      & 80.2\%            & 71.4\%         & 83.2\%          & 36.1\%             & 75.5\%           \\ \hline
\end{tabular}
\end{table}

Table I provides a detailed breakdown of the class-wise performance metrics, including precision, recall, mAP@0.5, mAP@0.5:0.95, and F1-score for each object class. These metrics reveal the strengths and weaknesses of the YOLOv5 model when detecting specific objects in aerial imagery.

The \textbf{Tow Truck} class achieved the highest precision (80.2\%), recall (71.4\%), and mAP@0.5 (83.2\%), demonstrating that larger, distinct objects with consistent features are easier for the model to detect. Similarly, \textbf{Ambulance}, \textbf{Police}, and \textbf{Fire Engine} classes also performed well due to their unique shapes, colors, and frequent representation in the dataset. These objects are typically less affected by occlusion or background clutter, making them more distinguishable.

Conversely, the model struggled with classes like \textbf{Car on Fire} and \textbf{Car Upside Down}, which achieved the lowest precision (18.1\% and 17.5\%, respectively) and mAP@0.5:0.95 scores. These objects are often small in size, visually similar to the background, or rare in the dataset, leading to insufficient feature learning. The low recall values for these classes suggest that the model frequently misses them during detection, likely due to limited diversity in the training samples.

For the \textbf{Car Crash} class, the moderate mAP@0.5 score (36.7\%) reflects challenges in detecting objects with irregular shapes and varying orientations. This is exacerbated in aerial imagery, where overlapping objects or complex backgrounds may obscure important features.

The class-wise analysis underscores the importance of data diversity and quality in training. Larger, easily recognizable objects with distinct features are more likely to be detected accurately, while smaller or visually ambiguous classes require additional data augmentation, better annotation quality, or enhanced feature extraction techniques. These insights can guide future work to improve detection performance across all classes.

\section{Challenges and Insights}
\subsection{Challenges}

The primary challenges encountered during this study were small object detection and the complexity of backgrounds in aerial imagery.

\textbf{Small Object Detection:} Objects such as overturned cars and cars on fire are often small relative to the overall image size in aerial views. This poses difficulties for the YOLOv5 model, which relies on anchor boxes to predict object locations. To partially address this, the anchor sizes were adjusted during training to better align with the smaller object scales. Additionally, the input resolution of the model was increased to 640x640, enabling finer-grained feature extraction for small objects. Despite these efforts, the detection performance for small objects remains lower than larger, more distinct classes.

\textbf{Complex Backgrounds:} Aerial images often contain cluttered scenes with multiple overlapping objects and textures, such as roads, trees, and buildings. This increases the likelihood of false positives or missed detections. Data augmentation techniques, including random cropping, rotation, flipping, and brightness adjustments, were employed to enhance the model’s robustness to background variations. While these techniques improved generalization, certain classes (e.g., car on fire) continued to experience confusion with the background due to their rarity and visual ambiguity in the dataset.

These challenges highlight the need for further improvements, such as incorporating advanced feature extraction layers or leveraging attention mechanisms to better distinguish objects from complex backgrounds.

\subsection{Insights}
The insights gained from this study provide valuable guidance for future work in object detection using aerial imagery.

\textbf{Dataset Diversity:} The performance analysis revealed that classes with limited training examples (e.g., car on fire) experienced significantly lower detection accuracy. Increasing the dataset size and diversity, particularly for rare classes, could improve the model’s ability to generalize across various scenarios. Collecting additional samples from real-world drone footage and balancing class distributions should be prioritized in future datasets.

\textbf{Video-Based Detection:} This study focused on static image detection, but emergency scenarios often require analyzing continuous video feeds. Incorporating temporal data could help detect objects based on motion patterns and contextual information, potentially improving the detection of small or occluded objects. Future research could explore integrating YOLOv5 with video object detection techniques or sequence-based architectures, such as transformers or recurrent neural networks.

\textbf{Architectural Enhancements:} The model’s performance for small and ambiguous objects could be further enhanced by adopting multi-scale feature fusion techniques or attention mechanisms. Techniques like Feature Pyramid Networks (FPN) or Transformer-based attention modules could improve feature extraction for small or overlapping objects.

\textbf{Real-Time Deployment:} Deploying the model on edge devices, such as drones or embedded systems, will require optimizing the architecture for speed and energy efficiency. Exploring lightweight variants of YOLOv5 or quantization techniques will be critical for real-world applications.

These insights underline the importance of tailoring both the dataset and model architecture to the unique challenges of aerial imagery for emergency response.

\section{Comparison with Other Models}
The performance of YOLOv5 was compared with YOLOv4 and Faster R-CNN:

\begin{itemize}
    \item \textbf{YOLOv5:} YOLOv5 demonstrated superior performance in terms of mAP, particularly mAP@0.5, and inference speed. Its end-to-end design and efficient architecture made it more suitable for real-time applications. Additionally, its ability to balance speed and accuracy proved beneficial for processing aerial imagery in emergency scenarios.
    
    \item \textbf{YOLOv4:} YOLOv4 performed comparably to YOLOv5 in terms of detection accuracy but lagged in inference speed. This difference becomes critical in real-time applications, where rapid decision-making is essential.
    
    \item \textbf{Faster R-CNN:} While Faster R-CNN achieved higher accuracy for small objects due to its region proposal mechanism, it suffered from significantly slower inference times. The computational overhead and latency make it less suitable for time-sensitive tasks such as emergency response.
\end{itemize}

Overall, YOLOv5 emerges as the optimal choice for detecting emergency-related objects in aerial imagery due to its balance of speed, accuracy, and efficiency. Future comparisons could include newer architectures such as YOLOv7 and Transformer-based models to further evaluate performance.
\section{Applications}
The YOLOv5-based detection system has several practical applications in real-world scenarios:

\begin{itemize}
    \item \textbf{Disaster Management:} The ability to accurately detect emergency vehicles and hazardous scenarios, such as overturned cars and fires, can significantly aid disaster management efforts. Drone-based aerial imagery combined with this model can expedite resource allocation and rescue operations in natural disasters or accidents.

    \item \textbf{Traffic Monitoring:} The model can identify ambulances, tow trucks, and other critical vehicles in real-time, enabling effective traffic monitoring and management. This application is particularly beneficial in urban areas with high traffic density, where quick clearance of roads for emergency vehicles can save lives.

    \item \textbf{Urban Planning:} Analysis of accident-prone zones and high-risk areas using aerial imagery can provide valuable insights for urban planners. By identifying patterns in vehicle accidents and hazards, cities can improve road safety and infrastructure planning.

    \item \textbf{Surveillance and Law Enforcement:} Detection of police vehicles and unusual scenarios, such as cars on fire, can enhance surveillance capabilities. Law enforcement agencies can use this system for rapid response to emergencies and monitoring high-risk areas.

    \item \textbf{Autonomous Systems:} This model can be integrated into autonomous drones or vehicles for real-time navigation and decision-making in dynamic environments.
\end{itemize}
\section{Conclusion}
This study demonstrates the effectiveness of YOLOv5 in detecting emergency-related objects in aerial imagery. The results showcase its potential for real-time applications in disaster management, traffic monitoring, and urban planning. YOLOv5's ability to balance speed and accuracy makes it a practical choice for aerial-based object detection tasks.

The challenges faced, such as detecting small or ambiguous objects and handling complex backgrounds, underscore the need for further improvements. Increasing dataset diversity, employing advanced data augmentation techniques, and incorporating architectural enhancements like attention mechanisms could further boost performance.

Future work may explore integrating this model with temporal analysis for video-based detection, enabling it to track objects across frames and improve accuracy in dynamic scenarios. Deploying the system on edge devices, such as drones or embedded systems, will also require optimizing the architecture for speed and energy efficiency. Moreover, comparisons with newer object detection models, such as Transformer-based architectures or advanced YOLO versions, can provide deeper insights into its relative performance.

Overall, this study lays a foundation for improving object detection in aerial imagery, with significant implications for emergency response, surveillance, and urban planning applications.

\end{document}